# Q-learning optimization in a multi-agents system for image segmentation

Issam Qaffou*, Mohamed Sadgal, Abdelaziz Elfazziki

Département Informatique,
Faculté des Sciences Semlalia, Université Cadi Ayyad,
Bd Prince My Abdellah BP 2390,
Marrakech, Morocco
* i.qaffou@gmail.com
{sadgal, elfazziki}@ucam.ac.ma

*Abstract*--**To know which operators to apply and in which order, as well as attributing good values to their parameters is a challenge for users of computer vision. This paper proposes a solution to this problem as a multi-agent system modeled according to the Vowel approach and using the Q-learning algorithm to optimize its choice. An implementation is given to test and validate this method.**

*Keywords: Q-learning; Parameter adjustment; Operator selection; Segmentation; Multi-agents system.*

## I. Introduction

To accomplish a vision task, the user must have in-depth knowledge about the appropriate operators to apply and their parameters to adjust. Indeed, the result of an image processing depends on the operator applied and the values attributed to its parameters. Inexperienced users should try all available operators and all possible parameter values to find the ones that give the best result. The manual process is not easy even for simple cases involving the application of a single operator with a single parameter to adjust. The number of attempts a user must make equal the number of possible values to assign to the parameter. As the range of values is high, as the computational complexity becomes higher. If the operator has more than one operator to adjust, the user must build all possible combinations of parameter values and test them all to find the one that gives the best result. Thus, the treatment process becomes even more complicated with only one operator, but with several parameters to adjust. However, the real computer vision tasks are completed in phases, and from each phase the user must choose one operator to apply. The user must know the best operators for each phase and the best values of their parameters. Often in such cases, we resort to experts in the field of computer vision to optimize the tests to be performed.

Some solutions have been proposed to help the user in his choice. For example, Pandore [1] a standardized library of image processing operators, using Ariane as Visual Programming Environment to select operators from the proposed list, and then chain them to form graphs of dataflow. But to activate an operator, it provides the user with a set of parameters that must be set manually. Also, Draper proposed in 2000 a system of recognition of objects he called ADORE [2]. Although this system is based on a robust method theoretically, it cannot guarantee good results as it uses a very specific and limited library of operators. ADORE is not interested in the problem of parameters adjusting. Furthermore, other research has focused on this issue only. For example, Nickolay et al. used evolutionary algorithms to design a method optimizing automatically parameters of vision system operators [3]. Few years later, Taylor [4] proposed a reinforcement learning framework, which uses connectionist systems as approximators to manage the problem of determining the optimal parameters for an application of computer vision. Unfortunately, these proposals remain limited to a well-defined type of images in special circumstances part. Moreover, they do not provide general methods to choose initially operators to use and then adjust their settings. Thus the overall problem to solve in this paper is both the selection of "good" operators and also the optimal adjustment of their parameters.

To solve this problem, an automatic solution using the Q-learning algorithm in a multi agent system is proposed in this paper. This solution generates from a library of operators, all operational chain (combination) to use in order to achieve an advocated goal by satisfying some constraints Fig.1.

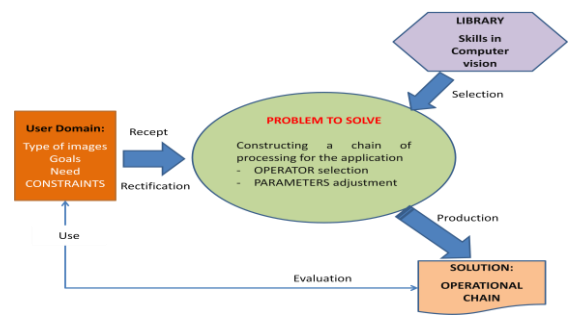

Fig.1. General schema of the generation of operator chains for an image application.

Thereafter, the control system of which the architecture is equipped should handle each combination by turning the optimal parameter values of each operator.

In the following two sections, an overview on the Q-learning algorithm and another on multi-agent system modeling are given respectively. Section 4 will detail the proposed solution. The implementation of this solution will be discussed in Section 5. Finally, Section 6 concludes the article.

## II. Q-learning

One of the most important discoveries in reinforcement learning was the development of an off-policy learning algorithm, known by Q-learning. This algorithm was proposed by Watkins in 1989 [5], and Sehad [6] worked on this approach to propose a model of the learning process Q-learning by highlighting the following three functions:

- **Selection function:** based on the knowledge stored in the internal memory, an action is selected and executed from the current situation as perceived by the system.
- **A Reinforcement function:** after the execution of the action in the real world, the reinforcement function uses the new situation to generate the reinforcement value.
- **Function update:** uses the reinforcement value to adjust the associated value with the state or the pair state, action that has been executed.

The simplest form of Q-learning, *1-step Q-learning* is defined as:

$Q(s_t, a_t) \leftarrow$
$(s_t, a_t) + \alpha[r_{t+1} + \gamma \max_a Q(s_{t+1}, a) - Q(s_t, a_t)]$

In this case, the value function of the learned action Q, approaches directly from Q* the optimal value function of the action, regardless of the policy. This greatly simplifies the analysis of the algorithm and provides early proofs of convergence.

The Q-learning algorithm is shown in a procedural form in Table 1.

| |
|---|
| Initialize $Q(s, a)$ arbitrarily |
| Repeat (for each episode): |
|     Initialize s |
|     Repeat (for each step of the episode): |
|         Choose an action a from s by using a policy derived from Q (e.g. $\varepsilon$-greedy) |
|         Take an action a, observe r, s' |
|         Choose an action a' from s' by using a policy derived from Q (e.g. $\varepsilon$-greedy) |
|         $Q(s_t, a_t) \leftarrow Q(s_t, a_t) + \alpha[r_{t+1} + \gamma \max_a Q(s_{t+1}, a) - Q(s_t, a_t)]$   (1) |
|         $s \leftarrow s'$; |
|     Until a terminal s |

**Table 1: Q-learning algorithm.**

The elements to consider in a Q-learning algorithm are multiple [7] [8]:

**Time**: The space of time has different forms; it can be discrete or continuous, finite or infinite, and fixed or random. Most studies on reinforcement learning use a discrete space-time.

**States**: they characterize situations of an agent and the environment at any moment; they can be divided into three forms:
- A relational position of the agent to the environment (position, etc.).
- A situation of the environment (environmental changes).
- An internal situation of the agent (his memory, sensors, etc.).

The three forms of state may be present at the same time depending on the problem being addressed.

**Actions**: an agent chooses an action among the possible actions at each time t; this action may be instantaneous or last until next time. Each state of the state space is associated with a set of possible actions from the action space.

**Reinforcement signal:** at each instant, the interaction produces a reinforcement value r, bounded numeric value, which measures the accuracy of the reaction agent. The purpose of the agent is to maximize the "accumulation" of these reinforcements in time.

Another important element of reinforcement learning is the policy that defines the behavior of the agent in a given time. It binds the visited states to the appropriate actions. Three types of policies are used in reinforcement learning: Boltzmann, $\varepsilon$-greedy, and greedy. The Boltzmann policy is a "softmax" method that uses the Gibbs distribution to estimate the probability of taking an action "a" in a given state "s". In the greedy policy, the agent selects the most interesting actions (as an evaluation function) in a given state. In this case, the agent does not explore all actions. In the $\varepsilon$-greedy policy the agent selects the greedy action with probability 1-$\varepsilon$ and other with probability $\varepsilon$. The $\varepsilon$-greedy policy is the most popular method for balancing the exploration and exploitation. Generally, the choice of appropriate policy depends on the application.

## III. Multi-agent system modeling

The theme of multi-agent systems (MAS), if it is not new, is currently a very active field of research. This discipline is the connection of several specific areas of artificial intelligence, distributed computer systems and software engineering. It is a discipline that focuses on collective behaviors produced by the interactions of several autonomous and flexible entities called agents.

This section introduces, first, the concepts of agents and multi-agent systems (MAS), then a vowel (A+E+I+O) modeling approach of a multi-agent system.

### A. Multi-agent system

Before dealing with multi-agent systems, it is useful to recall some basics about agents; it is the subject of the next subsection.

*1) Agent*

There are multiple definitions for an agent according to the privileged point of view.

In this work, the definition adopted by [9] according to M.Wooldridge's work is used: "An agent is a computer system situated in an environment, which acts in an autonomous and flexible way to meet the objectives for which it was developed".

The concepts of "situated", "autonomous" and "flexible" are defined as follows:
- situated: The environment provides agents with sensory inputs that allow them to act on it. In our context, the agent environment is the image.
- autonomous: the agent is able to act without the intervention of a third party (human or agent) and controls its own actions and its internal state.
- flexible: in this case the agent is:
  - able to respond in time: the agent must be able to perceive its environment and to develop a response within the required time;
  - proactive: an agent does not just react to its environment but it is also capable of producing himself actions motivated by goals.
  - social: the agent must be able to interact with other agents (software and human), when the situation requires it in order to complete tasks or help other agents to accomplish their own.

Of course, depending on the application, some properties are more important than others, it may even be that for certain types of applications, additional properties are required.

*2) Multi-agent system (MAS)*

A multi-agent system consists of a set of computer process occurring at the same time, so a set of multiple agents living at the same time, sharing common resources and communicating with each other. The key point in multi-agent systems is the formalization of the coordination between agents.

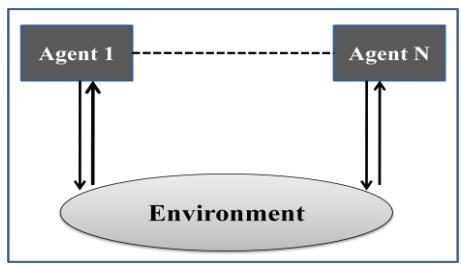

Fig. 2: A multi-agent system from an external observer point of view.

The agents are able to perceive and act on common shared environment, Fig.2. Perceptions enable agents to acquire information about the changes in their environment, and their actions allow them to change it.

*B. MAS modeling*

Several approaches exist for MAS modeling: O-MaSE, MESSAGE, PASSI, TROPOS, vowels, etc. This paper adopt the Vowel method for its simplicity, which consists of defining four components [10]:
- Agent: defining the internal structure and operation of each agent;
- Environment: determining the field of evolution and functioning of agents;
- Interaction: developing the ways by which agents communicate;
- Organization: global structure of the agents set.

The agent was already defined.

*1) Environment*

It is a common "space" to all agents in the system, which they perceive and interact with. It is the context where encoded rules in the agents, interactions and organizations are updated.

The environment can be accessible / inaccessible or deterministic / non-deterministic, episodic / non-episodic, continuous / discrete, static / dynamic, with or without rational opponent.

*2) Interaction*

Interaction is an important concept in multi-agent systems, because a goal from the definition of a society of agents is to create interaction between these agents to increase the capacity of the group. The interaction can be defined as dynamic linking of two or more agents through a set of reciprocal actions [11].

Interactions are based on:
- Cooperation: Working together for a common goal.
- Coordination: Organizing the solution of a problem so that harmful interactions are avoided or beneficial interactions are exploited.
- Negotiation: reaching an agreement acceptable to all concerned parties.

*3) Organization*

Organizations allow to master and structuring agents, who have just a partial view of the assembly in order to avoid inconsistencies and redundancies at the system level.

## VI. The proposed approach

To solve the problem of the choosing operators and adjusting their parameters automatically, a MAS architecture that uses Q-learning algorithm is proposed in this paper. This architecture is modeled according to the approach "Vowel" presented above. The four main components must be then defined for this method.

*A. Environment*

Given a vision task, the environment is defined as the images to be processed, their ground truths,

the agents constituting the MAS and any a priori knowledge. The proposed model generally proceeds by trial / error by comparing the result with the ground truth given by a domain expert.

## B. Agent

In the proposed architecture, we use three types of agents: a User Agent (UA), an Operator Agent (OA) and a Parameters Agent (PA). The agent UA is responsible for delivering all information necessary to perform the requested task. The agent OA is responsible for building all possible combinations of operators to apply. For each combination of operators, the agent OA generates a specialized agent AP to adjust the parameters of these operators.

### 1) Agent UA

The agent UA provides all possible phases from which will pass the global processing, and also their order. For each phase, it provides its applicable operators. Each operator has a set of parameters to be adjusted so that it can be applied; the agent UA proposes then the ranges of possible values for each parameter. The phase sequence determines the order of execution of operators.

The agent UA provides also the features extracted from each ground truth. This is a reference for comparison during reinforcement learning, Fig.3.

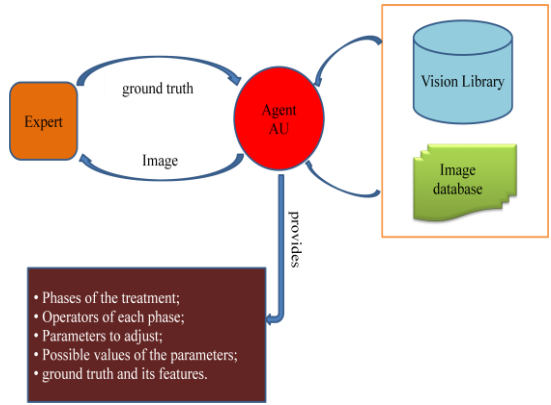

Fig. 3: Global schema of the agent UA

### 2) Agent OA

The agent OA builds all possible combinations according to the phases proposed by the agent UA. For each combination of operators, it generates an agent PA to adjust the parameters of these operators, Fig.4.

The agent OA generates as many agents PA as combinations built. After processing each combination by the AP agent by assigning the best values to the parameters of its operators, this combination is returned to the agent OA with its quality of processing. The combination having the highest quality is returned to the agent AU by the agent OA.

The quality of processing is determined by a comparison with the obtained result and the ground truth. It is important to notice that the selection of an operator as the best is not practical in most vision tasks requiring the execution of a chain of operators. Estimating an operator is best depends on all the phases of processing, that's why we talk about the best combination of operator and not the best operator.

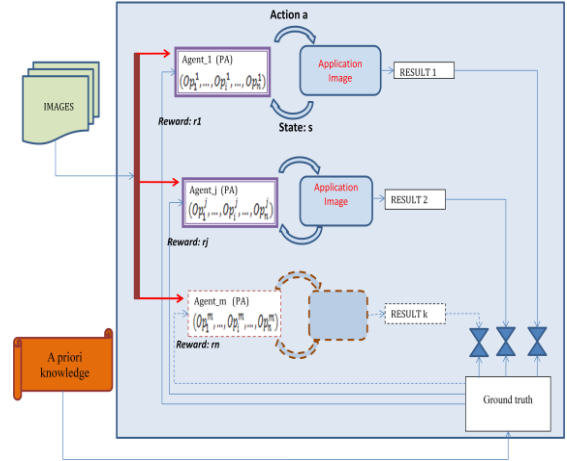

**Fig.4: to each operator combination is associated an agent PA.**

### 3) Agent AP

It is the agent responsible of processing the operator combinations constructed by the agent OA. He proceeds by reinforcement learning to find the best parameter values for each operator. For this purpose, he built all possible combinations of parameter values, to assign them to each operator combination involved in a test on the input image. Each combination of values gives a different result. To evaluate each combination of values, a comparison is made between the obtained results and the desired one (ground truth). Thus, an evaluation is produced for each combination, and the one having the maximum evaluation is adopted by the agent PA, and represents the final values to be assigned to the parameters of the operators in the combination in question.

The agent PA is a reinforcement learning agent using Q-learning algorithm in the optimization process. Thus, the agent PA must be adaptable to the reinforcement learning structure in terms of defining actions, states and reward. Fig.5 illustrates the general approach of the agent PA explained above.

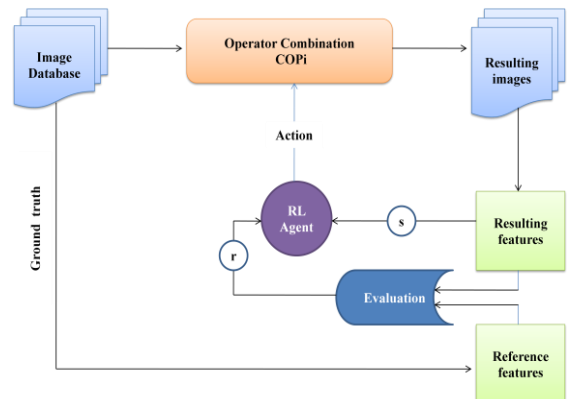

Fig.5: general schema of the functioning of the agent PA.

*a) State definition*

Generally, we define a state $s_i$ representing an image in an iteration I, by a function of a set of features $\chi_j$ :

$$s_i = f(\chi_1, \chi_2, \ldots \chi_n)$$

The features that can be used depend on the task at hand and the specifications of the problem to solve.

For example, for a segmentation task, the features can be:
- Features quantifying the number of contours in the images;
- Features that reflect global information at the pixel level (White pixels) in grayscale;
- Etc.

*b) Action definition*

Any modification of a value represents an action for the agent PA. The set of actions is the set of all possible combinations of parameter values.

Given a combination of operators: ($OP^1_i$, $OP^2_i$,…, $OP^n_i$).

Each operator $OP^j_i$ has a set of parameters:

$$(P^j_{i1}, P^j_{i2}, \ldots, P^j_{iq})$$

Each parameter $P^j_{ik}$ has a range of possible values: $V^j_{ik} = \{v^{j1}_{ik}, v^{j2}_{ik}, \ldots, v^{jm'}_{ik}\}$

An elementary action of the operator $OP^j_i$ is then: $a_z = (u^{jz}_{ik}, \ldots, u^{jz}_{ik})$ where $u^{jz}_{ik} \in V^j_{ik}$

Therefore, an action for the agent PA is defined by a combination of elementary actions defined above:

$$a = (a_1, a_2, \ldots, a_n)$$

*c) Return definition*

The return is usually defined for vision tasks, according to a quality criterion representing if an image has been well processed or not. The return could be a reward if the agent PA chooses a good action, if not it is a punishment. A simple method to calculate the return is to compare the obtained result with the ground truth (image processed manually by an expert). The similarity between the resulting image and the reference image determines the quality of processing, and therefore evaluates the selected action. In the context of this paper, for a segmentation using contour approaches we calculate error measures that give global clues on the quality of a result: error over-detection, error under-detection, error localization, etc.. While for segmentation using region approaches, we calculate for example, Yasnoff errors [12] or Vinet criterion [12], etc. After calculating the similarity criteria, we assess the results obtained by using a weighted sum D of scalar differences of these criteria.

$$D = \sum_i w_i D_i$$

The weights $w_i$ are chosen according to the importance of each criterion $D_i$.

In the experience section, the definition of actions, States and Return will be concretized on a segmentation application.

## C. Interaction

To allow passage between the agents of the proposed architecture, interactions between the three pairs agents are developed. The agent UA interacts with the agent OA by providing him all the necessary information in order to return the best combination of operators to apply. The agent OA interacts with the agent PA by separating him the combinations of operators in order to return each combination with the best values of its parameters. Thus, the agent UA interacts with the agent PA through the agent OA.

## D. Organization

The three agents UA, OA and PA are organized in the proposed architecture as shown in Fig.6.

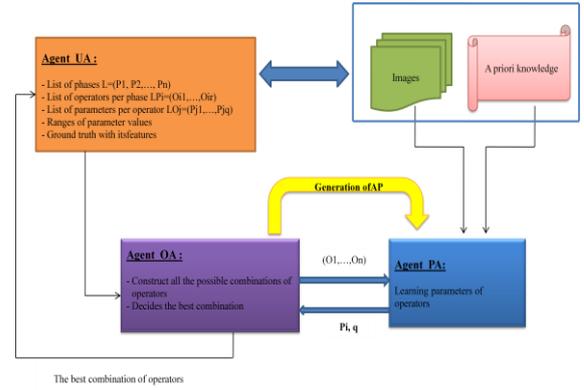

Fig.6. Global structure of the three agents.

## V. Experience

The proposed approach is tested on an application of image segmentation using contour approaches. The goal is not giving a new method of segmentation, but from a set of possible solutions, which one is the best?

The experiment uses a database containing 100 different images. Each image has its ground truth, which is an image manually segmented by an expert. The approach is implemented in Matlab. How the three types of agents are defined?

## A. Agent UA

He identifies three phases of processing with the operators of each one: pre-processing, processing and post-processing.

*Pre-processing*: this phase is to improve the quality of image by using filters. The agent UA proposes three operators in this phase **'medfilt2'**, **'ordfilt2'** and **'wiener2'** with one parameter to be adjusted for each: the size of the filter, with two possible values are: 3 or 5.

*Processing*: In this phase the contours contained in the image are detected. The agent UA proposes a single operator for this phase: **'edge'**, with two parameters to adjust. The first is to choose the filter (sobel, Prewitt zerocross, log) and the second concerns the threshold to eliminate low-contrast edges [0.02,..., 0.09, 0.1].

*Post-processing:* this phase consists of refining the image by removing small objects. The agent UA proposes a single operator for this phase **'bwareaopen'** with two parameters to adjust: the first defines the maximum size (in pixels) of objects to be deleted. The second defines the connectivity: the number of neighbors to consider 4 or 8. For 2D images the connectivity by default is 8.

### B. Agent OA

The agent OA builds all possible combinations of operators:
$COP_1$= (medfilt2, edge, bwareaopen)
$COP_2$= (wiener2, edge, bwareaopen)
$COP_3$= (ordfilt2, edge, bwareaopen)

The question is which one of these $COP_i$ is the best to apply?

For each combination, an agent PA is generated to find the optimal values for the parameters of its operators.

### C. Agent PA

The process of parameters adjustment is done by the Q-learning algorithm which requires the definition of actions, states and return function.

*Actions*: They are all possible combinations of parameter values. We choose another action by selecting other parameter values. An example of an action: Action= [3, ('sobel', 0.02), (5,8)]

*States*: A state is defined as a set of features extracted from the image. In our application a state is defined by three features:
$$s = [\chi_1, \chi_2, \chi_3]$$
$\chi_1$ is the ratio between the number of calculated contours and the number of the contour in the reference;
$\chi_2$ is the ratio between the number of white pixels obtained and their number in the reference;
$\chi_3$ is the ration between the length of the longest contour resulted and the length of the longest one in the reference.

*Return*: it is defined by a weighted sum of three error measures that give global clues on the quality of a segmentation result by contour: error of over-detection, error of under-detection and error localization [30]. These criteria evaluate the result of boundary detection. A weighted sum D is used to enhance measure considered effective in maximizing return. The weights are also selected in this direction.

$$D = \omega_1 D_1 + \omega_2 D_2 + \omega_3 D_3$$

The agent PA works on each combination of operators following the schema in Fig.5. The proposed image database constitutes the environment of the agent AP. The images are closely similar in their features in order to not change the environment for every test. The goal is to know which one of the combinations built by the agent OA gives the best segmentation. Each image has three results of segmentation. Fig.7 contains five images taken randomly from the process of treatment and the best segmentation for each of them. These results are given by the same combination of operators which is **COP2 = (wiener2, edge, bwareaopen)** having respectively the parameters: **3 ; (prewitt , 0.02) ; (5 , 8).**

So this combination is returned to the agent UA by the agent OA as the best sequence of operator to apply for segmenting such images. The obtained result depends essentially on the used images, the proposed operators and the range of possible values of parameters. For images more complicated, the library of operator should be well defined.

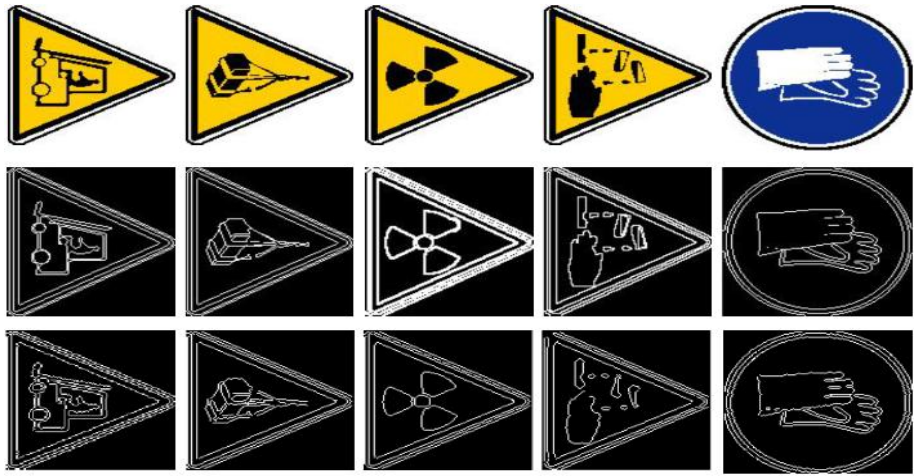

Fig.7: from top to bottom: original images, the ground truth and the result of our approach.

## IV. Conclusion

In this paper a new method to solve the problem of operator selection and parameter adjustment for a computer vision task is presented. The proposed solution is based on robust theoretical techniques, Q-learning and MAS, which allow it to give good results. It was tested in a segmentation task, in order to find between several combinations of operators which one is the best to use. The obtained results reinforce the theoretical concept of the method. As perspective, the ambition is to build a system which could give for each type of images the appropriate combination of operators to use for segmentation, and on another way a system which could classify new images in its corresponding type. This is the subject of our next work.

# References

1. R. Clouard, A. Elmoataz & F.Angot, "*PANDORE : une bibliothèque et un environnement de programmation d'opérateurs de traitement d'images*", Rapport interne du GREYC, Caen, France, Mars 1997.
2. B.A. Draper, J. Bins, and K. Baek, "*ADORE: Adaptive Object Recognition*". Videre, 2000. 1(4): p. 86-99.
3. B. Nickolay, B. Schneider, S.Jacob, "*Parameter Optimization of an Image Processing System using Evolutionary Algorithms*" 637-644. CAIP 1997.
4. G. W.Taylor, "*A Reinforcement Learning Framework for Parameter Control in Computer Vision Applications*" Proceedings of the First Canadian Conference on Computer and Robot Vision (CRV'04), IEEE 2004.
5. C. J. C. H. Watkins. "*Learning from Delayed Rewards*". PhD thesis, Cambridge University, 1989.
6. S. Sehad, "*Contribution à l'étude et au développement de modèles connexionnistes a apprentissage par renforcement : application a d'acquisition de comportements adaptatifs*". Thèse génie informatique et traitement du signal. Montpellier : Université de Montpellier II, 1996, 112 p.
7. O. Buffet, "*Apprentissage par Renforcement dans un système multi-agents*" : rapport de stage DEA. DEA informatique. Nancy : UFR STEMIA, université Henri Poincaré-Nancy 1, 2000, 40p.
8. O. Buffet, "*Une double approche modulaire de l'apprentissage par renforcement pour des agents intelligents adaptatifs*". Thèse UFR STEMIA. Nancy : Université Henri Poincaré-Nancy 1, 2003, 215p.
9. R Haroun. "*Segmentation des tissus cérébraux sur des images par résonance magnétique*". Master's thesis, Université des sciences et de la technologie Houari Boumediène, 2005.
10. J. Simao, Y. Demazeau: "*On Social Reasoning in Multi-Agent Systems*". Inteligencia Artificial, Revista Iberoamericana de Inteligencia Artificial 5(13): 68-84, 2001.
11. J. Ferber. "*Les Systèmes Multi-Agents Vers une intelligence collective*". Inter-Editions, 1995.
12. DO Minh Chau, "*Évaluation de la segmentation d'images*". Rapport final TIPE. Institut de la francophonie pour l'informatique. Nanoï 2007.